\documentclass{article}

\usepackage{arxiv}

\usepackage[utf8]{inputenc} 
\usepackage[T1]{fontenc}    
\usepackage{hyperref}       
\usepackage{url}            
\usepackage{booktabs}       
\usepackage{amsfonts}       
\usepackage{nicefrac}       
\usepackage{microtype}      
\usepackage{lipsum}
\usepackage{graphicx}

\title{BET: A Backtranslation Approach for Easy Data Augmentation in Transformer-based Paraphrase Identification Context}

\author{Jean-Philippe Corbeil \\
  Polytechnique Montreal \\
  \small\texttt{jean-philippe.corbeil@polymtl.ca} \\\And
  Hadi Abdi Ghadivel \\
  Polytechnique Montreal \\
  \small\texttt{hadi.abdi-ghavidel@polymtl.ca} \\}

\begin{document}
\maketitle

\begin{abstract}
Newly-introduced deep learning architectures, namely BERT, XLNet, RoBERTa and ALBERT, have been proved to be robust on several NLP tasks. However, the datasets trained on these architectures are fixed in terms of size and generalizability. To relieve this issue, we apply one of the most inexpensive solutions to update these datasets. We call this approach BET by which we analyze the backtranslation data augmentation on the transformer-based architectures. Using the Google Translate API with ten intermediary languages from ten different language families, we externally evaluate the results in the context of automatic paraphrase identification in a transformer-based framework. Our findings suggest that BET improves the paraphrase identification performance on the Microsoft Research Paraphrase Corpus (MRPC) to more than 3\% on both accuracy and F1 score. We also analyze the augmentation in the low-data regime with downsampled versions of MRPC, Twitter Paraphrase Corpus (TPC) and Quora Question Pairs. In many low-data cases, we observe a switch from a failing model on the test set to reasonable performances. The results demonstrate that BET is a highly promising data augmentation technique: to push the current state-of-the-art of existing datasets and to bootstrap the utilization of deep learning architectures in the low-data regime of a hundred samples.    
\end{abstract}


\section{Introduction}

Machine learning and deep learning algorithms have achieved impressive results lately. A part of this success is due to the availability of a large amount of annotated data. The majority of the public NLP datasets lack a large amount of data, which limits the accuracy of the models. On the other hand, the provision of big data is costly and time-consuming. In this paper, we intend to increase the size of natural language data through an easy data augmentation technique called BET. 

Data augmentation has been well-established in computer vision tasks (e.g. Shorten and Khoshgoftaar \cite{shorten2019survey}), but it is not a widespread practice in the NLP community. In the past couple of years, data augmentation in NLP has gained growing interest. According to Wei and Zou \cite{wei2019eda}, it is costly and challenging to increase the size of the textual data. Thus, fewer efforts have been seen in the state-of-the-art (SOTA). In this paper, we used the backtranslation approach, one of the most successful methods for phrase-based translation \cite{bojar2011improving,sennrich2016improving,lample2017unsupervised}. In other words, we consider the backtranslation technique to act as a paraphraser and we evaluate the augmented data on the four transformer-based architectures: BERT, XLNet, RoBERTa and ALBERT.

\newpage

Our main contributions are as follows:
\begin{itemize}
    \item We systematically divided the Google Translate (GT) languages into family clusters and select up to ten languages as intermediary languages.
    \item We augmented using BET the whole MRPC paraphrase corpus, as well as a downsampled version of Quora \footnote{https://www.kaggle.com/c/quora-question-pairs} and Twitter Paraphrase Corpus \cite{lan2017continuously}, which we released for reproducibility\footnote{\url{https://github.com/jpcorb20/wikipedia-lang-families}}\footnote{ \url{https://github.com/jpcorb20/google-translate-backtranslation-da}}\footnote{ \url{https://github.com/jpcorb20/bet-backtranslation-paraphrase-experiment}}.
    \item Using the augmented data, we analyzed the improvement in terms of precision, recall, F1-score and accuracy for four transformer-based models.
\end{itemize}

The structure of the current paper is as follows: In section 2, we review the most recent works on data augmentation and then discuss what has been done in the paraphrase identification task, especially on MRPC Paraphrase Corpus, Quora Question Duplicate and Twitter Paraphrase Corpus. In section 3, we explain our data, baselines and methodology. In section 4, we describe our results and discuss them in detail. The last section ends the paper in our conclusion, limitations and future work.

\section{Previous Works}
\label{intro}
\subsection{Data Augmentation}

Data augmentation in NLP has been an active research area mainly since 2018. In this section, we highlight what methods have been used and how they had an impact on NLP tasks. In what follows, we categorize the data augmentation techniques into linguistic and non-linguistic categories. By linguistic property, we mean the meaning is preserved after the data augmentation and the augmented data follows the correct linguistic form.  

In the linguistic category, the augmentation is done based on word-level and sentence-level replacement or an entire generation of new sentences. In the word-level replacement, the words are usually replaced with the following alternatives: Replacing with the synonyms in the thesaurus \cite{zhang2015character,wei2019eda,rizos2019augment} and Replacing with the words of most similar embedding vectors \cite{wang2015s, wolfe2019data,giridhara2019study}. Another type of word-level is to substitute words with the masked words by a suggestion from a BERT model (e.g. CBERT \cite{wu2019conditional}). This type of augmentation is enhanced by Reinforcement learning in works like \cite{hu2019learning,niu2019automatically}.

In the sentence level replacement, the alternative sentence is generated through the following methods: Backtranslation \cite{tong2019supervised,coulombe2018text,sennrich2016improving}, Paraphrase generation with regex \cite{coulombe2018text} and XLDA technique which selects at random a segment and translates it \cite{singh2019xlda}.

Entire-generation of new sentences is mostly done by the transformers based on the classification of the labels. Anaby-Tavor et al. \cite{anaby2019not} used the GPT-2 model \cite{radford2019language} to generate new examples and filter them by using a classifier trained on the original data. In a similar work, Kumar et al. \cite{kumar2020data} combined GPT-2 \cite{radford2019language}, BERT \cite{vaswani2017attention} and BART \cite{lewis2019bart}, and then elaborated a conditional data augmentation framework by prepending the class labels to text sequences.

In the non-linguistic category, the augmentation is conducted through the methods such as:
\begin{itemize}
    \item Random insertion \cite{wei2019eda}
    \item Random swap \cite{wei2019eda}
    \item Random deletion \cite{wei2019eda}
    \item Textual noise injection \cite{coulombe2018text}
    \item Spelling error \cite{coulombe2018text}
\end{itemize}

The closest work to ours in terms of the task and datasets belongs to Shakeel et al. \cite{shakeel2020multi}. The authors used LSTM and CNN with hand-crafted features on paraphrase corpora: Quora, MRPC and SemEval. They achieved the results competitive with the SOTA by augmenting the paraphrasing data with a graph-based technique on the syntax tree. Nevertheless, the current SOTA results from transformer-based architectures are beyond their reported results. Therefore, there is a need to verify the impact of the backtranslation data augmentation technique, particularly in the transformer-based architectures.

\subsection{Paraphrase Identification on MRPC}


 In this section, we highlight how automatic paraphrase identification has recently been trained on the accessible MRPC dataset.
 
 A large-scale labelled corpus of MRPC introduced by Dolan and Brockett \cite{dolan2005automatically} is collected from news data. Among several models trained on this corpus, transformer-based models like BERT were the most successful ones. 
 

Mainly before 2018, most of the paraphrase identification tasks on MRPC were conducted through using hand-crafted features such as sentence similarities metrics like Cosine distance, L2 Euclidean distance, word embeddings as well as part-of-speech tag embeddings \cite{he2015multi}, syntactic and semantic features \cite{filice2015structural}, enhanced embeddings by adding syntactic and multi-sense features \cite{cheng2015syntax}, and combination of latent features with fine-grained n-gram overlap\cite{ji2013discriminative}.

Based on the attention-based transformer model (Vaswani et al. \cite{vaswani2017attention}), BERT \cite{devlin2018bert} made a revolutionary change in the NLP SOTA in many GLUE tasks, especially paraphrase identification. Both accuracy and F1 score increased tremendously by BERT. However, the BERT pretraining step has some drawbacks which were addressed on the following modified versions. In RoBERTa, Liu et al. \cite{liu2019roberta} removed the next-sentence pretraining objective to make improvements on the BERT masked language modelling objective. The dependency between words became visible by XLNet \cite{yang2019xlnet}, which is pretrained based on the potential permutations of context words surrounding a target word. Followed by XLNet, ALBERT \cite{lan2019albert} was introduced to address the issues of high memory consumption and BERT's training speed. Among the tasks performed by ALBERT, paraphrase identification accuracy is better than several other models like RoBERTa. Research on how to improve BERT is still an active area, and the number of new versions is still growing. The examples are StructBERT \cite{wang2019structbert}, ERNIE \cite{sun2019ernie2}, etc. Overall, the paraphrase identification performance on MRPC becomes stronger in newer frameworks. 

Although we have seen the significant improvement in the paraphrase identification task in the past years due to the introduction of the transformer-based models, the data augmentation effect on them has less been touched. As a result, we aim to figure out how carrying out the augmentation influences the paraphrase identification task performed by these transformer-based models. 

\section{Methodology}
\subsection{Data}
In this paper, our first experiment is on the full MRPC corpus. This corpus is introduced inside the GLUE benchmark (General Language Understanding Evaluation) \cite{wang2019glue} for NLP language models and seems to be one of the most known corpora in the paraphrase identification task. This corpus consists of sentence pairs which are automatically extracted from online news sources. We call the first sentence "sentence" and the second one, "paraphrase". There are 4076 pairs for the train and 1725 for the test set. Overall, there are 5801 pairs in the original set provided by Microsoft. 

In our second experiment, we analyze the data-augmentation on the downsampled versions of MRPC and two other corpora for the paraphrase identification task, namely the TPC and Quora dataset. Lan et al. \cite{lan2017continuously} published TPC as a new version of the SemEval-2015 task 1 on "Paraphrase and Semantic Similarity in Twitter" \cite{xu2014twitter,xu2014datatwitter}. 6 Amazon Mechanical Turkers annotated it with a good correlation with expert's annotations. When four or more Turkers are agreeing on the similarity, the pair is considered a paraphrase. When two or less agree, it is a non-paraphrase pair. Finally, the pairs with three Turkers agreeing and disagreeing are removed, because of their ambiguity. The Quora Question Duplicate dataset was released as a Kaggle Competition in 2017.

\subsection{Downsampling Procedure}

We performed downsampling on our three datasets in a balanced fashion by which we consider the equal number of positive and negative examples. We aimed to show the impact of our BET approach in a low-data regime and generalize it to other corpora within the paraphrase identification context. Previous authors used this type of approach \cite{wei2019eda,giridhara2019study,anaby2019not}. In this regard, 50 samples are randomly chosen from the paraphrase pairs and 50 samples from the non-paraphrase pairs. This selection is made in each dataset to form a downsampled version with a total of 100 samples.

\subsection{Translation Setup}
\begin{figure*}[ht!]
   \centering
   \includegraphics[width=\linewidth]{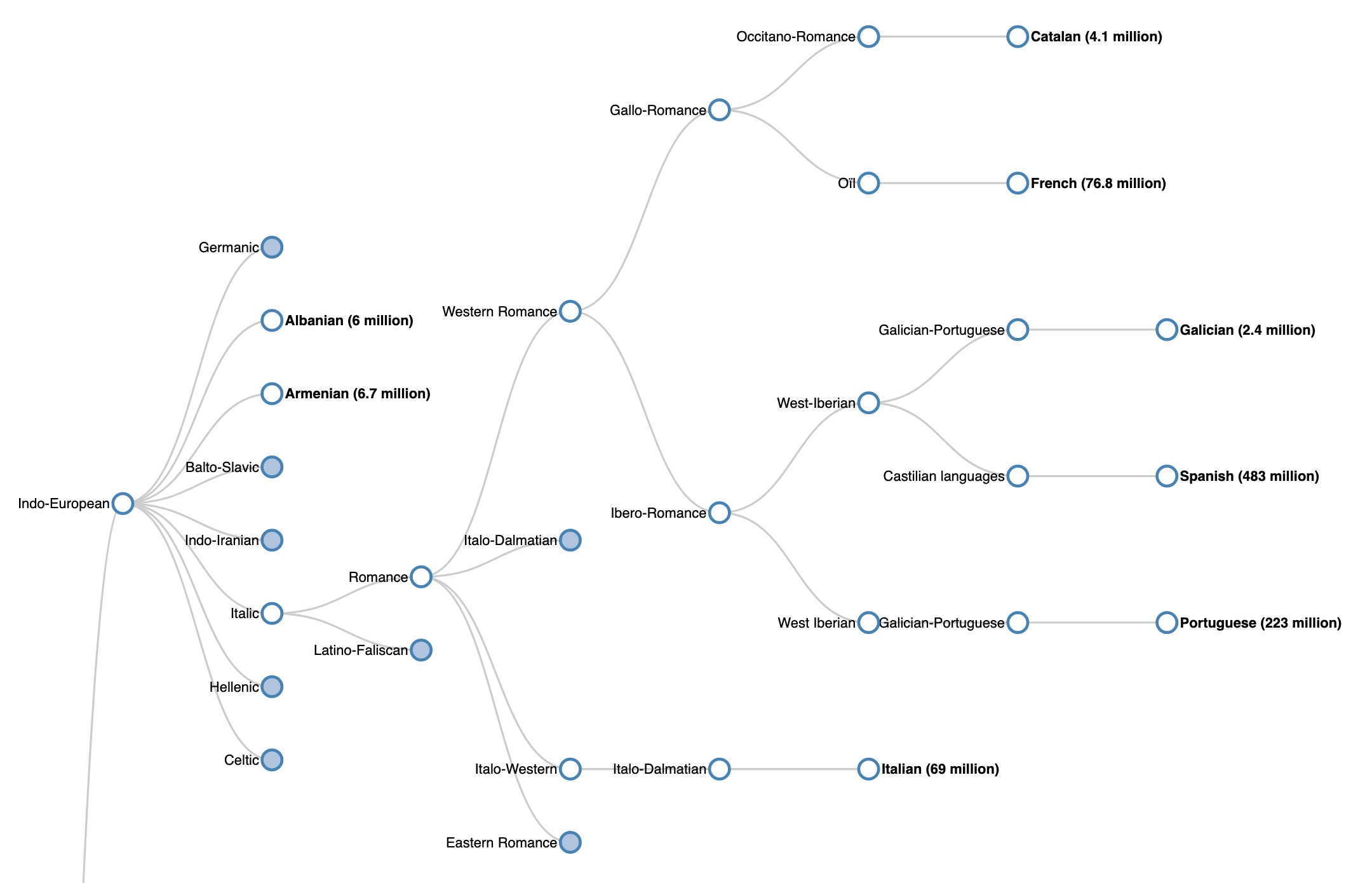}
   \caption{Romance branch in the language family tree extracted from Wikipedia info-boxes.}
   \label{fig:romance_lang_tree}
\end{figure*}

The Google Translate (GT) API can translate 109 languages into each other. Some of these languages fall into family branches, and some others like Basque are language isolates. We systematically selected ten languages based on the following procedure:

\begin{enumerate}
    \item We clustered all the languages into the related language families based on the information provided in the Wikipedia info-boxes\footnote{Code and full interactive version available at \url{https://github.com/jpcorb20/wikipedia-lang-families}}. The Romance branch is illustrated for instance in Figure \ref{fig:romance_lang_tree}.
    \item Based on the maximum number of L1 speakers, we selected one language from each language family. With this process, we aimed at maximizing the linguistic differences as well as having a fair coverage in our translation process. This led us to Table \ref{tab:top_lang}.
    \item We kept the top-10 languages.
\end{enumerate}

\begin{table}[ht!]
  \centering
  \fontsize{10}{10}
  \caption{Top-10 Languages sorted by most L1 speakers for each language family.}
    \begin{tabular}{|c|c|c|}
    \hline
    \textbf{Language} & \textbf{Family} & \textbf{Native} \\
    (GT code)& & \textbf{speakers} \\
    & & (in million) \\
    \hline
     \textbf{Chinese} (zh)& Sino-Tibetan & 1,200 \\
    (Simplified) & & \\
    \hline
    \textbf{Spanish} (es)& Indo-European & 483 \\
    \hline
    \textbf{Arabic} (ar)& Afro-Asiatic & 310 \\
    \hline
    \textbf{Japanese} (ja)& Japonic & 125\\
    \hline
    \textbf{Telugu} (te)& Dravidian & 82 \\
    \hline
    \textbf{Javanese} (jv)& Austronesian & 82 \\
    \hline
    \textbf{Korean} (ko)& Koreanic & 77.2 \\
    \hline
    \textbf{Vietnamese} (vi)& Austroasiatic & 76 \\
    \hline
    \textbf{Turkish} (tr)& Turkic & 75.7\\
    \hline
    \textbf{Yoruba} (yo)& Niger-Congo & 40 \\
    \hline
    \end{tabular}
  \label{tab:top_lang}
\end{table}


\subsection{Paraphrase Identification}
Our main goal is to analyze the data-augmentation effect on the transformer-based architectures. As a result, we select BERT \cite{devlin2018bert}, RoBERTa  \cite{liu2019roberta}, XLNet \cite{yang2019xlnet} and ALBERT \cite{lan2019albert}. As discussed in Section 2, subsequent models modified BERT to improve it as much as possible. 

As the quality in the paraphrase identification dataset is based on a nominal scale ("0" or "1"), paraphrase identification is considered as a supervised classification task. We input the sentence, the paraphrase and the quality into our candidate models and train classifiers for the identification task.

\subsubsection{Data Augmentation}

In the current study, we aim to augment the paraphrase of the pairs and keep the \emph{sentence} as it is. Therefore, our input to the translation module is the paraphrase. Once translated into the target language, the data is then back-translated into the source language. Our filtering module removes the backtranslated texts, which are an exact match of the original paraphrase. Overall, our augmented dataset size is about ten times higher than the original MRPC size, with each language generating 3,839 to 4,051 new samples. For instance, one paraphrase pair from MRPC was:

\begin{itemize}
    \item \textit{Sentence}: They had published an advertisement on the Internet on June 10, offering the cargo for sale, he added.
    \item \textit{Paraphrase}: On June 10, the ship's owners had published an advertisement on the Internet, offering the explosives for sale.
\end{itemize}

When using BET with Chinese (zh) as intermediary language, we received the new paraphrase: 

\begin{itemize}
    \item \textit{Augmented paraphrase}: On June 10, the ship owner posted an advertisement on the Internet to sell explosives.
\end{itemize}

\subsection{Experimental Setup}
\label{section3.5}

We conducted our experiments with the HuggingFace library \cite{wolf2019transformers} and ran the experiments locally on a NVIDIA RTX2070 GPU, making our results easily reproducible. Our hyperparameters are tuned based on what is proposed by \cite{wolf2019transformers} for the BERT (\textit{BERT base uncased}), RoBERTa model (\textit{RoBERTa base}) and ALBERT (\textit{ALBERT base v2}) models, which is as follows:

\begin{itemize}
    \item Epochs = 3 
    \item Batch size = 32
    \item Learning rate = 3e-5
    \item Maximum sequence length = 128
    \item random seed = 42
\end{itemize}

With XLNet (\textit{XLNet base cased}), we had to change the batch size to 16.

Overall, our experiment were conducted along two categorical sets: the set of models $\Sigma$ and the set of intermediary languages $L$. More formally, we define them as $L=\{zh,es,ar,ja,te,jv,ko,vi,tr,yo\}$ and $\Sigma=\{bert,xlnet,roberta,albert\}$. The dataset set $D$ is further added, which is $D=\{mrpc,tpc,quora\}$. For each model and dataset pairs in $\Sigma \times D$, we evaluated a baseline (\textit{base}) to compare all our results obtained with the augmented datasets. We also computed results for the augmentation with all the intermediary languages (\textit{all}) at once.

The performance of the paraphrase identifier is evaluated based on the following set $M$ of metrics: 
\begin{itemize}
    \item \textbf{Accuracy (Acc)}: Proportion of correctly identified paraphrases.
    \item \textbf{Precision (P) (positive predictive value)}: Proportion of correctly identified paraphrases among all the predicted paraphrases.
    \item \textbf{Recall (R) (sensitivity)}:
    Proportion of correctly identified paraphrases among all the existing paraphrases.
    \item \textbf{F1 score (F1):} The harmonic mean of precision and recall.
\end{itemize}

In our visualizations, we looked specifically at the gain $g_m$ in these different metrics $m \in M$ which is defined as:

\begin{equation}
    g_{m}(\sigma,d,l) = m(\sigma,d,l) - m(\sigma,d, base)
\end{equation}

Where $d \in D$ is a dataset, $l \in L$ is an intermediary language used for the augmentation, and $\sigma \in \Sigma$ is a model. To get a better overview of the efficiency of BET, we visualize the marginal gain distributions --- $G_m(d)$, $G_m(l)$ and $G_m(\sigma)$ --- along each of these categorical sets one at a time with each metric $m$, taking the other sets as free variables. For instance, we can have $G_m(\sigma)$ for all models in $\Sigma$, of which we can analyze the obtained gain by model for all metrics.

We divided the train set, full or downsampled, of all datasets into the train set and development set (20\%). All the results are computed on their respective test sets except for the Quora Question Duplicate dataset from which we split our own test set (20\%).

\section{Results and Discussion}
In this section, we discuss the results we obtained through training the transformer-based models on the original and augmented full and downsampled datasets. Also, we highlight which languages help boost the models.
\subsection{Full MRPC}

\begin{figure}[ht!]
   \centering
   \includegraphics[width=0.6\linewidth]{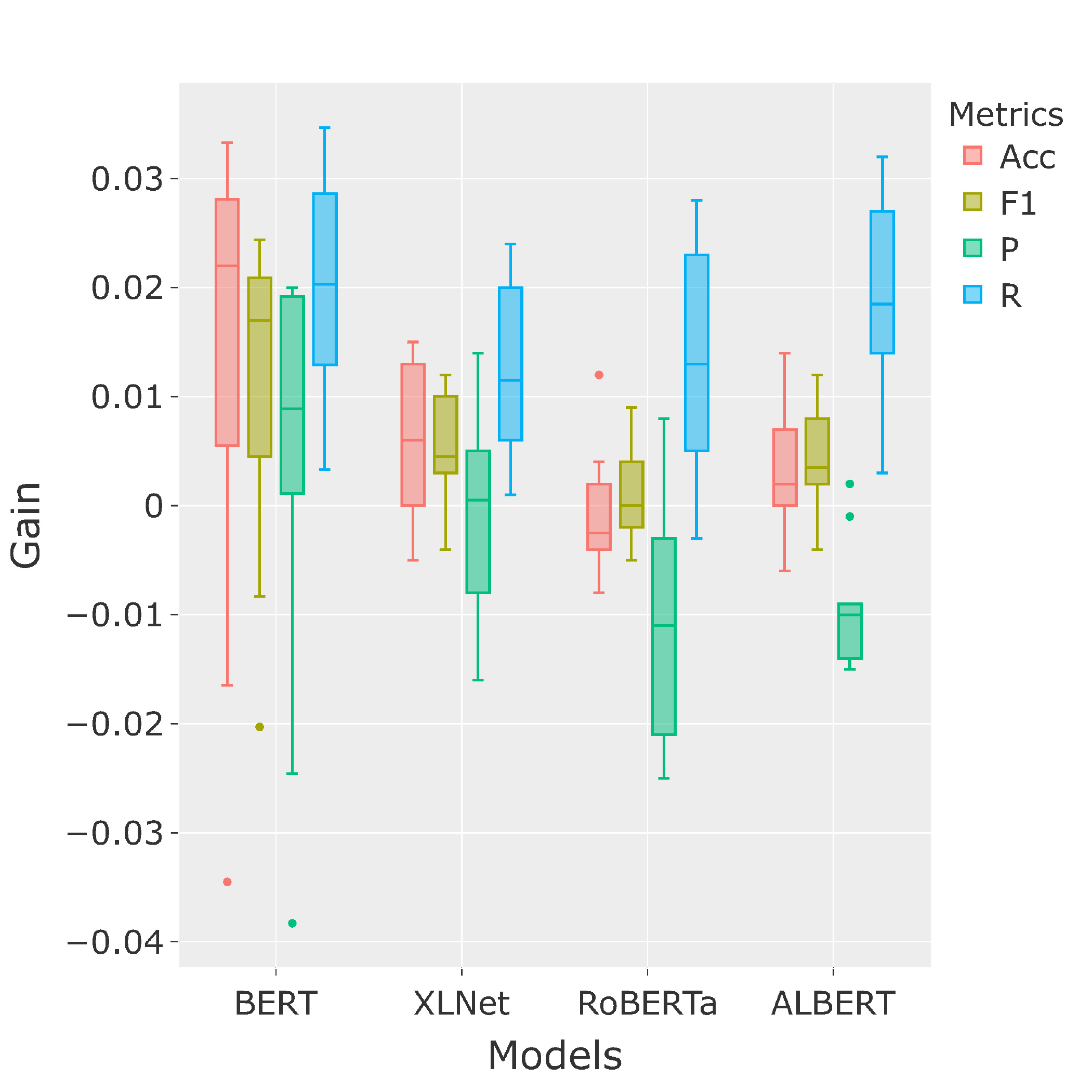}
   \caption{Gain $G_m(\sigma)$ for MRPC for each model.}
   \label{fig:model_results}
\end{figure}

Table \ref{tab:results} shows the performance of each model trained on original corpus (baseline) and augmented corpus produced by all and top-performing languages. As the table depicts, the results both on the original MRPC and the augmented MRPC are different in terms of accuracy and F1 score by at least 2 percent points on BERT. For the other models, the improvements range from 1 to 1.5 percent points. These observation are visible in Figure \ref{fig:model_results}. For precision and recall, we see a drop in precision except for BERT. The recall improvements are the highest gains for all models, approximately 3 percent points, except for XLNet. RoBERTa that obtained the best baseline is the hardest to improve while there is a boost for the lower performing models like BERT and XLNet to a fair degree. Nevertheless, we enhanced the SOTA performance of RoBERTa from an F1 score of 0.909 \cite{liu2019roberta} up to 0.915. This boosting is achieved through the Vietnamese intermediary language's augmentation, which leads to an increase in precision and recall. 

\begin{table}[htbp]
  \centering
  \small
  \caption{MRPC results with \textit{baseline} for all models as well as augmentation for \textit{all} and top-performing language.}
    \begin{tabular}{llllll}
    \hline
    \textbf{Model} & \textbf{Data} & \textbf{Acc} & \textbf{F1} & \textbf{P} & \textbf{R} \\
    \hline
   BERT & baseline & 0.802 & 0.858 & 0.820  & 0.899 \\
   BERT & all & 0.824 & 0.877 & 0.819 & \textbf{0.945} \\
   BERT & es & \textbf{0.835} & \textbf{0.882} & \textbf{0.840} & 0.929 \\
    \hline
    XLNet & baseline & 0.845 & 0.886 & 0.868 & 0.905 \\
    XLNet & all & 0.837 & 0.883 & 0.840 & \textbf{0.932} \\
    XLNet & ja & \textbf{0.860} & \textbf{0.897} & \textbf{0.877} & 0.919 \\
    \hline
    RoBERTa & baseline & 0.874 & 0.906 & 0.898 & 0.914 \\
    RoBERTa & all & 0.872 & 0.907 & 0.877 & \textbf{0.939} \\
    RoBERTa & vi& \textbf{0.886} & \textbf{0.915} & \textbf{0.906} & 0.925 \\
    \hline
    ALBERT & baseline & 0.853 & 0.890 & \textbf{0.885} & 0.895 \\
    ALBERT & all & 0.841 & 0.886 & 0.847 & \textbf{0.929} \\
    ALBERT & yo & \textbf{0.867} & \textbf{0.902} & 0.884 & 0.922 \\
    \hline
    \end{tabular}%
  \label{tab:results}%
\end{table}%

\begin{figure*}[ht!]
   \centering
   \includegraphics[width=\linewidth]{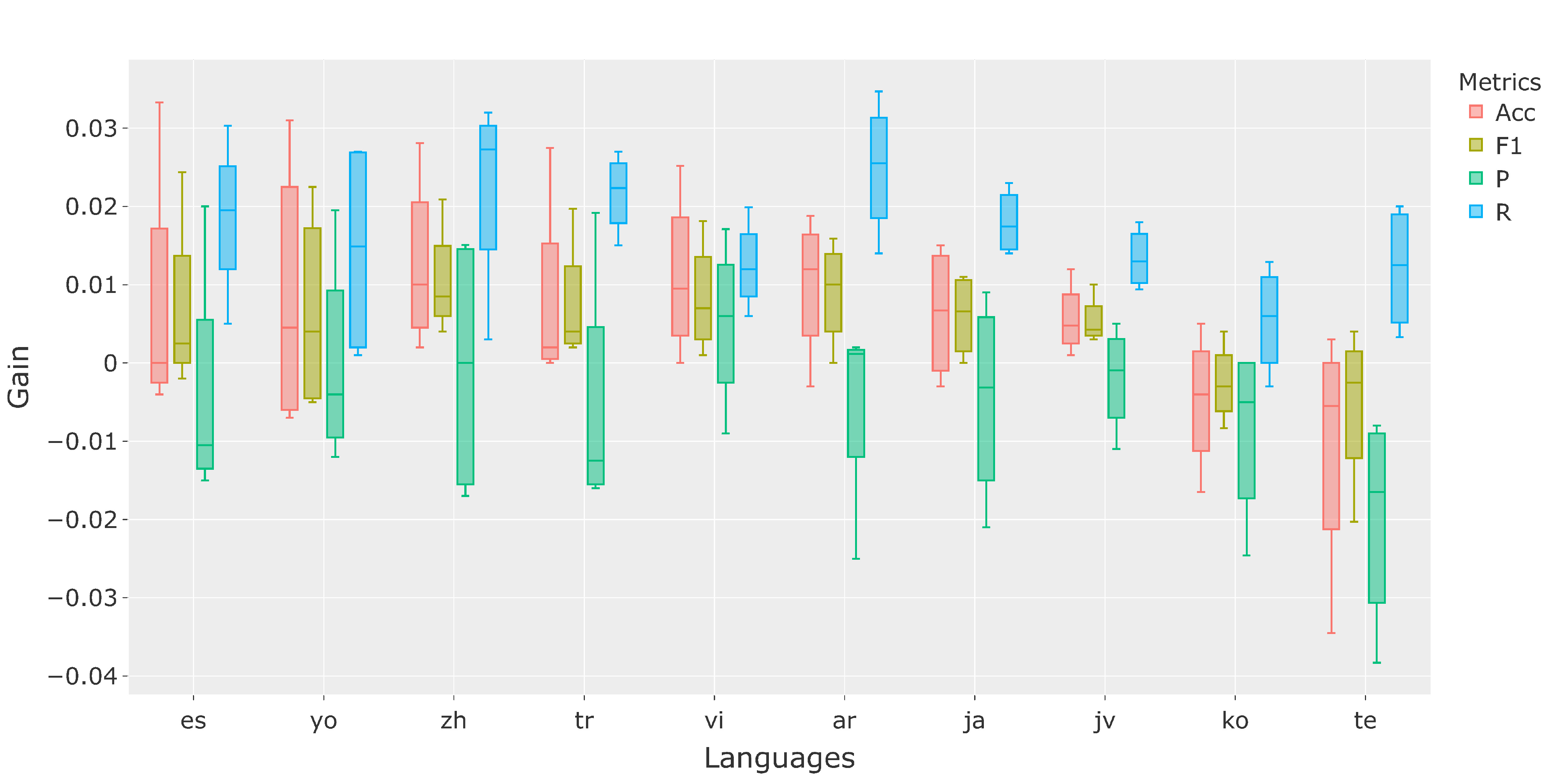}
   \caption{Gain $G_m(l)$ for MRPC for each intermediary language sorted by maximum accuracy.}
   \label{fig:lang_results}
\end{figure*}

The results for the augmentation based on a single language are presented in Figure \ref{fig:lang_results}. We improved the baseline in all the languages except with the Korean (ko) and the Telugu (te) as intermediary languages. Interestingly, we have the best results in terms of gains with Spanish-based augmentation (es) than the results based on all the language augmentations (see Table \ref{tab:results}). 

Overall, we see a trade-off between precision and recall. The first is dropping while the latter is rising. We already expected this phenomenon according to our initial studies on the nature of backtranslation in the BET approach. We trade the preciseness of the original samples with a mix of these samples and the augmented ones. This mixture led to a significant gain in the recall, but sometimes a drop in precision. On average, we observed an acceptable performance gain with the Arabic (ar), Chinese (zh) and Vietnamese (vi). We note that the best improvements are obtained with Spanish (es) and Yoruba (yo). In general, we hypothesize that the method can be used with this set of five intermediary languages.

\subsection{Downsampled MRPC, TPC and Quora}

\begin{figure*}[ht!]
   \centering
   \includegraphics[width=\linewidth]{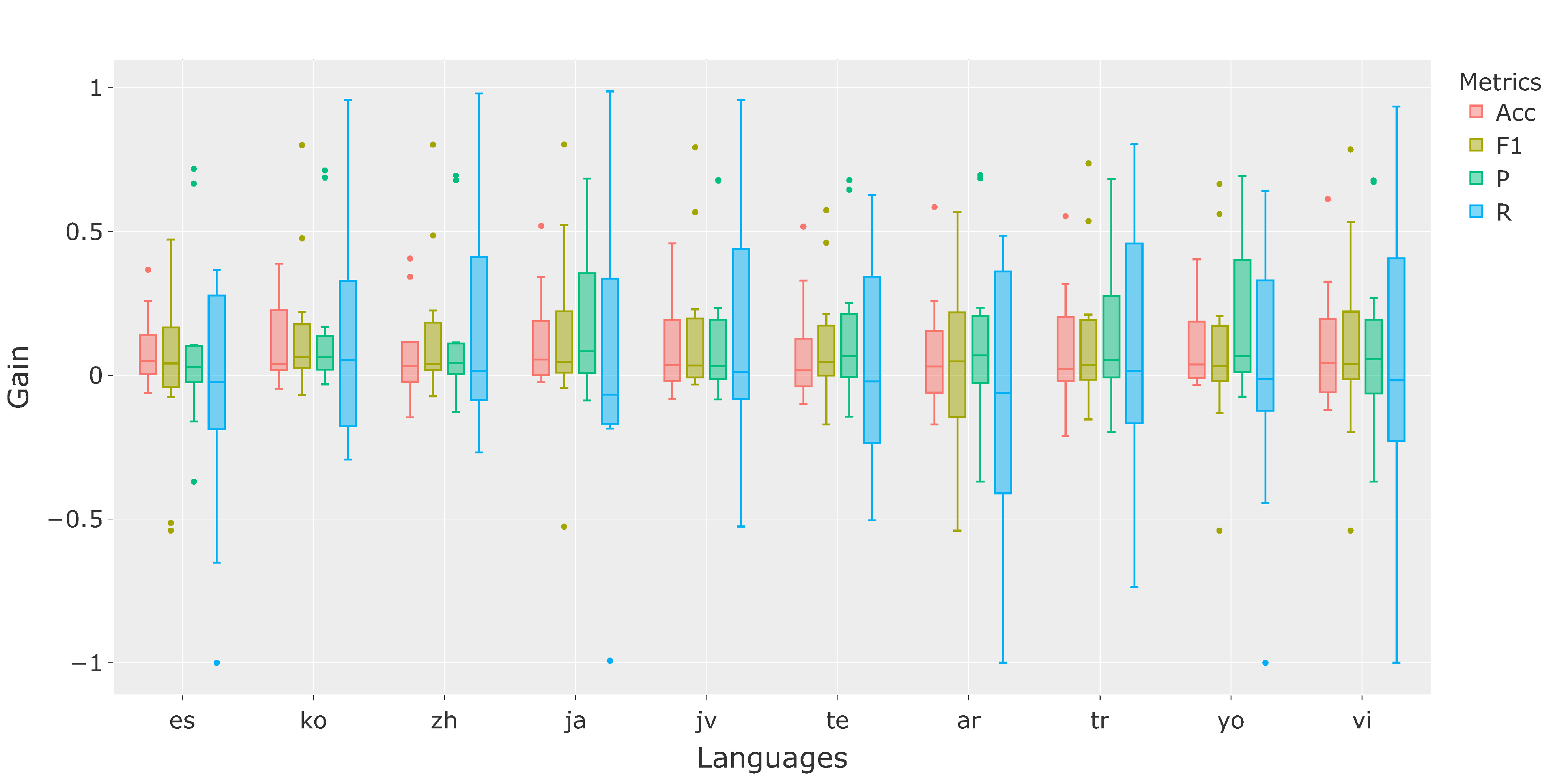}
   \caption{Gain $G_m(l)$ for each language with the downsampled datasets.}
   \label{fig:downsampled_lang_results}
\end{figure*}

We display the best F1 score results for the downsampled datasets of a 100 balanced samples in Tables \ref{tab:downsampled_results_mrpc}, \ref{tab:downsampled_results_tpc} and \ref{tab:downsampled_quora_results}. We found that many poor-performing baselines received a boost with BET. For the downsampled MRPC, the augmented data did not work well on XLNet and RoBERTa, leading to a reduction in performance. Nevertheless, the results for BERT and ALBERT seem highly promising. For TPC, as well as the Quora dataset, we found significant improvements for all the models.

\begin{table}[htbp]
  \centering
  \small
  \caption{Top F1 score results with their baseline (100 balanced samples) for our downsampled MRPC.}
    \begin{tabular}{llllll}
    \hline
    \textbf{Model} & \textbf{Aug.} & \textbf{Acc} & \textbf{F1} & \textbf{P} & \textbf{R} \\
    \hline
    BERT  & base  & 0.335 & 0.000 & 0.000 & 0.000 \\
    BERT  & ja & 0.677 & 0.802 & 0.676 & \textbf{0.987} \\
    \hline
    ALBERT & base & 0.677 & 0.774 & \textbf{0.722} & 0.834\\
    ALBERT & ja & \textbf{0.696} & \textbf{0.804} & 0.703 & 0.939\\
    \hline
    \end{tabular}%
  \label{tab:downsampled_results_mrpc}%
\end{table}%

\begin{table}[htbp]
  \centering
  \small
  \caption{Top F1 score results with their baseline (100 balanced samples) for our downsampled TPC.}
    \begin{tabular}{llllll}
    \hline
    \textbf{Model} & \textbf{Aug.} & \textbf{Acc} & \textbf{F1} & \textbf{P} & \textbf{R} \\
    \hline
    BERT  & base & 0.813 & 0.000 & 0.000 & 0.000\\
    BERT  & te & \textbf{0.862} & 0.574 & \textbf{0.645} & 0.517\\
    \hline
    XLNet & base & 0.803 & 0.426 & 0.447 & 0.407 \\
    XLNet & jv & 0.778 & 0.564 & 0.436 & 0.798\\
    \hline
    ALBERT & base & 0.734 & 0.535 & 0.391 & 0.850\\
    ALBERT & ja & 0.795 & \textbf{0.586} & 0.461 & 0.805\\
    \hline
    RoBERTa & base & 0.180 & 0.305 & 0.180 & \textbf{1.000}\\
    RoBERTa & vi & 0.793 & 0.536 & 0.449 & 0.667\\
    \hline
    \end{tabular}%
  \label{tab:downsampled_results_tpc}%
\end{table}%

\begin{table}[htbp]
  \centering
  \small
  \caption{Top F1 score results with their baseline (100 balanced samples) for our downsampled Quora.}
    \begin{tabular}{llllll}
    \hline
    \textbf{Model} & \textbf{Aug.} & \textbf{Acc} & \textbf{F1} & \textbf{P} & \textbf{R}\\
    \hline
    BERT  & base & 0.561 & 0.524 & 0.436 & 0.658\\
    BERT  & tr & 0.573 & 0.585 & 0.455 & 0.819\\
    \hline
    XLNet & base & 0.670 & 0.401 & \textbf{0.602} & 0.300\\
    XLNet & all & \textbf{0.713} & 0.661 & 0.585 & 0.761\\
    \hline
    ALBERT & base & 0.515 & 0.601 & 0.431 & 0.992\\
    ALBERT & ar & 0.648 & 0.649 & 0.512 & 0.885\\
    \hline
    RoBERTa & base & 0.374 & 0.540 & 0.370 & \textbf{1.000}\\
    RoBERTa & all & 0.691 & \textbf{0.668} & 0.553 & 0.843\\
    \hline
    \end{tabular}%
  \label{tab:downsampled_quora_results}%
\end{table}%

\begin{figure}[ht!]
   \centering
   \includegraphics[width=0.6\linewidth]{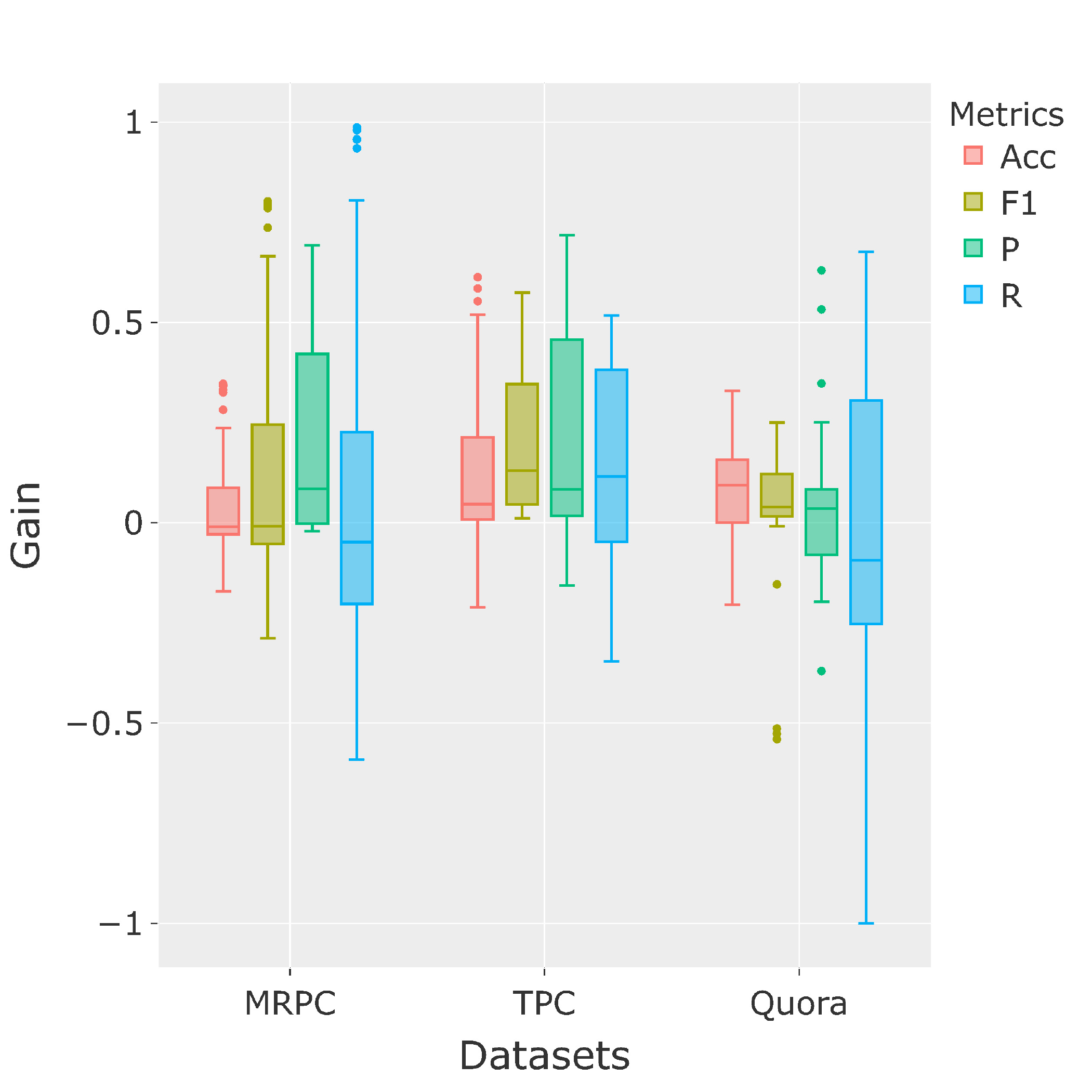}
   \caption{Gain $G_m(d)$ for each downsampled dataset.}
   \label{fig:downsampled_dataset_results}
\end{figure}

In Figure \ref{fig:downsampled_dataset_results}, we display the marginal gain distributions by augmented datasets. The downsampled TPC dataset was the one that improves the baseline the most, followed by the downsampled Quora dataset. For the Quora dataset, we also note a large dispersion on the recall gains. We explain this fact by the reduction in the recall of RoBERTa and ALBERT (see Table \ref{tab:downsampled_quora_results}) while XLNet and BERT obtained drastic augmentations. The augmentation of the downsampled MRPC did not work well. With Table \ref{tab:downsampled_results_mrpc}, we see only a considerable improvement for the BERT model, which is illustrated by the outliers in the top of Figure \ref{fig:downsampled_dataset_results}. 

\begin{figure}[ht!]
   \centering
   \includegraphics[width=0.6\linewidth]{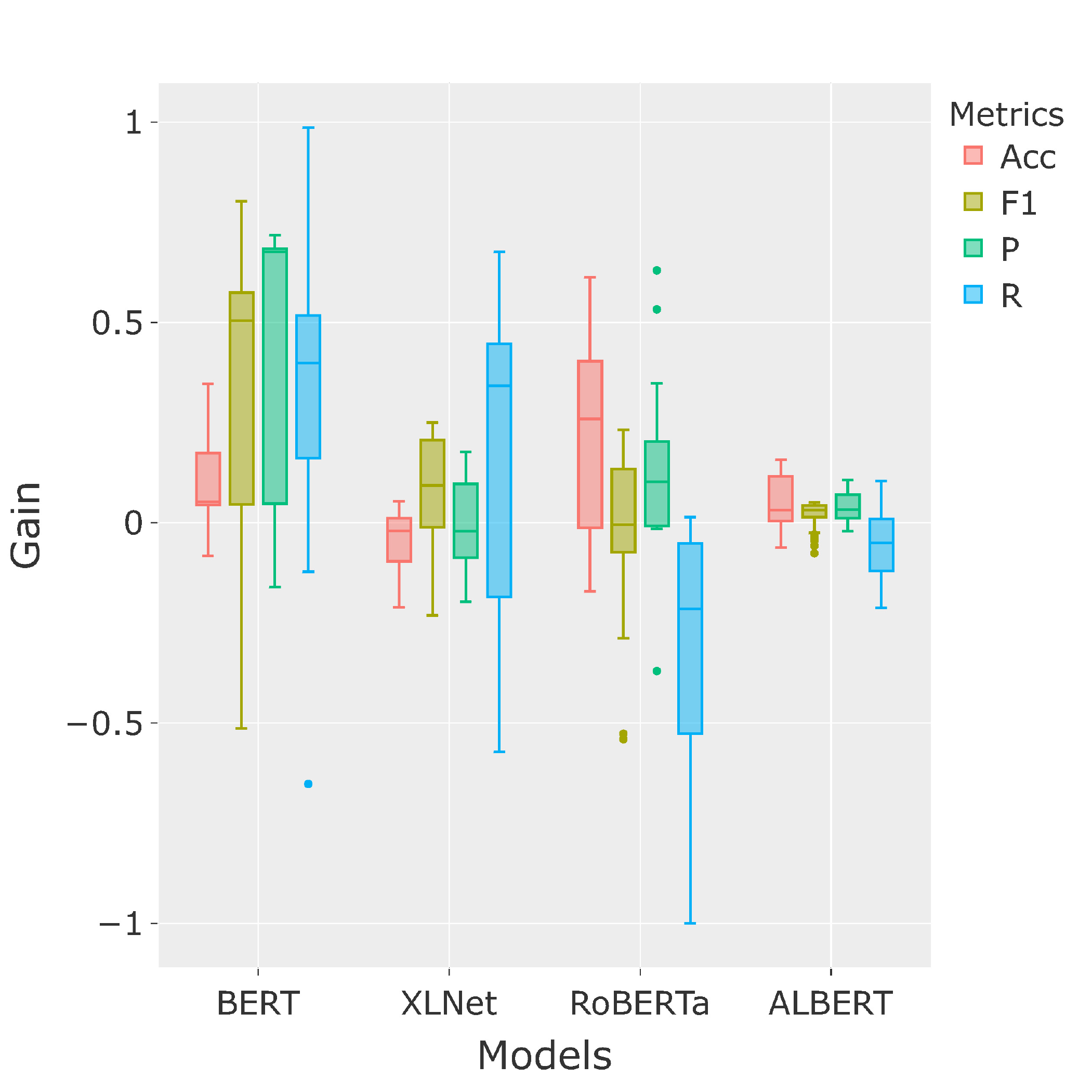}
   \caption{Gain $G_m(\sigma)$ for each model with the downsampled datasets.}
   \label{fig:downsampled_model_results}
\end{figure}

In Figure \ref{fig:downsampled_dataset_results}, we displayed the marginal gain distributions by augmented datasets. The downsampled TPC dataset was the one that improves the baseline the most, followed by the downsampled Quora dataset. For the Quora dataset, we also note a large dispersion on the recall gains. We explain this fact by the reduction in the recall of RoBERTa and ALBERT (see Table \r
When we consider the models in Figure \ref{fig:downsampled_model_results}, BERT improves the baseline considerably, explained by failing baselines of 0 as the F1 score for MRPC and TPC. XLNet had nearly no improvement in the accuracy, but the F1 score benefited from a 0.1 gain with high recall gain on average and small precision loss. RoBERTa gained a lot on accuracy on average (near 0.25). However, it loses the most on recall while gaining precision. On average, this results in no gain in F1 score. Finally, ALBERT gained the less among all models, but our results suggest that its behaviour is almost stable from the start in the low-data regime.

In Figure \ref{fig:downsampled_lang_results}, we also analyzed the marginal gains by languages. We noted a gain across most of the metrics. Our findings suggest that all languages are to some extent efficient in a low-data regime of a hundred samples. This motivates using a set of intermediary languages and not only rely on one language. By doing so, we can benefit from peculiar generated samples that maximize the performances when having a specific combination of data, model and intermediary language.

\section{Conclusion and Future Work}

In this paper, we described how we augmented the MRPC paraphrase corpus as well as the downsampled version of it besides TPC dataset and Quora Question Duplicate dataset. We showed that augmenting data through BET led to a gain in the performance metrics (accuracy, F1 score, precision and recall) for most transformer-based models, particularly BERT. We, further, showed that BET helps boost the paraphrase identification task in the SOTA. Among the other models, we achieved better results for the full MRPC corpus augmented with the Vietnamese language for the RoBERTa model in terms of the F1 score. In the low-data regime, we demonstrated reasonable performance gains up to large ones. By applying BET on three train sets --- MRPC, TPC and Quora Question Duplicate --- of only a hundred balanced samples each, we augmented drastically the number of available samples and observed a huge gain in performances on their respective test sets.

In conclusion, it seems we can find a combination of a dataset, model and intermediary language that can lead to a substantial gain in the performances. Thus, BET is guaranteed to be successful through using a set of intermediary languages like Arabic, Chinese, Vietnamese, Spanish and Yoruba. 

Despite the success we achieved in using BET for textual data augmentation, the current study has limitations which we plan to address in our future works:
\begin{itemize}
    \item We did not compare BET with previous techniques from the literature. Our future work will investigate how the datasets augmented by these techniques help advance the performance of transformer-based architectures.
    \item The experiments can be extended to other NLP tasks like automatic short answer grading, open question answering, etc.
    \item Since we filtered the back-translated data based on the exact match, we plan to check them based on the grammatical correctness and topic preservation.
    \item Using other deep learning architectures with the augmented datasets is also part of our future works.
\end{itemize}

\bibliographystyle{unsrt}  

\end{document}